\documentclass[11pt,a4paper]{article}

\usepackage[utf8x]{inputenc} % allow utf-8 input
\usepackage[T1]{fontenc}    % use 8-bit T1 fonts
\usepackage{lmodern}
\usepackage{url}            % simple URL typesetting
\usepackage{booktabs}       % professional-quality tables
\usepackage{amsfonts}       % blackboard math symbols
\usepackage{nicefrac}       % compact symbols for 1/2, etc.
\usepackage{microtype}      % microtypography
\usepackage{amsmath, amsthm, amssymb}
\usepackage{subcaption}
\usepackage{graphicx}
\usepackage{booktabs}

\usepackage{mathtools}
\usepackage{enumitem}

\newtheorem{proposition}{Proposition}

\usepackage{xcolor}

\let\olddelta\delta
\renewcommand{\delta}{\olddelta \hspace{-0.3mm}}

\newcommand{\deq}{\mathrel{\mathop:}=}
\renewcommand{\epsilon}{\varepsilon}
\newcommand{\target}{o^\ast}
\newcommand\trans{^\top}
\newcommand{\real}{\mathbb{R}}
\newcommand{\E}{\mathbb{E}}
\renewcommand{\Pr}{\mathbb{P}}

\title{Unbiasing Truncated Backpropagation Through Time}
\author{Corentin Tallec, Yann Ollivier}
\date{}

\begin{document}

\maketitle

\begin{abstract} \emph{Truncated Backpropagation Through Time} (truncated
    BPTT, \cite{jaeger2002tutorial}) is a widespread method for
    learning recurrent computational graphs.
    Truncated BPTT keeps the computational benefits of
    \emph{Backpropagation Through Time} (BPTT \cite{werbos:bptt}) while
    relieving the need for a complete backtrack through the whole data
    sequence at every step.  However, truncation favors short-term
    dependencies: the gradient estimate of truncated
    BPTT is biased, so that it does not benefit from the convergence
    guarantees from stochastic gradient theory. We introduce \emph{Anticipated Reweighted
    Truncated Backpropagation} (ARTBP), an algorithm that keeps the
    computational benefits of truncated BPTT, while providing
    unbiasedness. ARTBP works by using variable truncation lengths
    together with carefully chosen compensation factors in the
    backpropagation equation. We check the viability of ARTBP on two
    tasks. First,
a simple synthetic task where careful balancing of temporal dependencies at different scales is needed: truncated BPTT displays unreliable performance,
    and in worst case scenarios, divergence, while ARTBP converges
    reliably.
    Second, on Penn Treebank character-level language modelling \cite{ptb_proc},
    ARTBP slightly outperforms truncated BPTT.
\end{abstract}

%\NDY{Notation: Use $L$ for sequence length and $T$ for truncation? Or use
%$T_{\mathrm{max}}$ for total length? [might be ugly in
%$\mathcal{L}_T$]}

\emph{Backpropagation Through Time} (BPTT) \cite{werbos:bptt} is the de
facto standard for training recurrent neural networks. However, BPTT has
shortcomings when it comes to learning from very long sequences: learning a
recurrent network with BPTT requires unfolding the network through time for as
many timesteps as there are in the sequence. For long sequences this
represents a
heavy computational and memory load. This shortcoming is often overcome heuristically, by
arbitrarily splitting the initial sequence into subsequences, and
only backpropagating on the subsequences. The resulting algorithm is often
referred to as \emph{Truncated Backpropagation Through Time} (truncated
BPTT, see for instance
\cite{jaeger2002tutorial}).  This comes at the cost of losing long term
dependencies.

We introduce \emph{Anticipated Reweighted Truncated BackPropagation} (ARTBP), a
variation of truncated BPTT designed to provide an unbiased gradient estimate,
accounting for long term dependencies.  Like truncated BPTT, ARTBP splits the
initial training sequence into subsequences, and only backpropagates on those
subsequences. However, unlike truncated BPTT, ARTBP splits the training
sequence into variable size subsequences, and suitably modifies the
backpropagation equation to obtain unbiased gradients.

Unbiasedness of gradient estimates is the key property that provides
convergence to a local optimum in stochastic gradient descent procedures.
Stochastic gradient descent with biased estimates, such as the one provided by
truncated BPTT, can lead to divergence even in simple situations and even with
large truncation lengths (Fig.~\ref{fig:bpttf}).

ARTBP is experimentally compared to truncated BPTT. On truncated BPTT
failure cases, typically when balancing of temporal dependencies is key,
ARTBP achieves reliable convergence thanks to unbiasedness. On small-scale
but real world data, ARTBP slightly outperforms truncated BPTT on the test case
we examined.

ARTBP formalizes the idea that, on a day-to-day basis, we can perform
short term optimization, but must reflect on long-term effects once in a
while; ARTBP turns this into a provably unbiased overall gradient estimate.
Notably, the many short subsequences allow for quick
adaptation to the data, while preserving overall balance.

% Section~\ref{sec:back} covers the recurrent learning framework, where both
% ARTBP and truncated BPTT apply. ARTBP is formally described and proven under restrictions
% in Section~\ref{sec:artbp}. Experimental results are given in
% Section~\ref{sec:exp}.\NDY{Is this useful??}

\section{Related Work}
BPTT \cite{werbos:bptt} and its truncated counterpart \cite{jaeger2002tutorial}
are nearly uncontested in the recurrent learning field. Nevertheless, BPTT is
hardly applicable to very long training sequences, as it requires storing and
backpropagating through a network with as many layers as there are timesteps
\cite{ilya-thesis}. Storage issues can be partially addressed as in
\cite{constant-mem-bptt}, but at an increased
computational cost.  Backpropagating through very long sequences also
implies
performing fewer gradient descent steps, which significantly slows down learning
\cite{ilya-thesis}. 

Truncated BPTT heuristically solves BPTT deficiencies by chopping the initial
sequence into evenly sized subsequences. Truncated BPTT truncates gradient
flows between contiguous subsequences, but maintains the recurrent hidden state
of the network.  Truncation biases gradients, removing any theoretical
convergence guarantee.
Intuitively, truncated BPTT
has trouble learning dependencies above the range of truncation.  
\footnote{
    Still, as the hidden recurrent state is not reset between
    subsequences, it may 
    contain hidden information about the distant past, which can be exploited
    \cite{ilya-thesis}.
}

\emph{NoBackTrack} \cite{nobacktrack} and \emph{Unbiased Online Recurrent
Optimization} (UORO) \cite{uoro} both scalably provide unbiased online
recurrent learning algorithms. They take the more extreme point of view
of requiring memorylessness, thus forbidding truncation schemes and any
storage of past states. NoBackTrack and UORO's
fully online, streaming structure comes at the price of noise injection
into the gradient estimates via a random rank-one reduction. ARTBP's
approach to unbiasedness is radically different:  ARTBP
is not memoryless but does not inject artificial noise into the
gradients, instead, compensating for the truncations directly inside the
backpropagation equation.

\section{Background on recurrent models}
\label{sec:back}
The goal of recurrent learning algorithms is to optimize
a parametric dynamical system, so that its output sequence, or
predictions, is as close as possible to some target sequence, known a
priori. Formally, given a dynamical system with 
state $s$, inputs $x$, parameter $\theta$, and transition function $F$,
\begin{equation}
    s_{t+1} = F(x_{t+1}, s_t, \theta)
\end{equation}
the aim is to find a $\theta$ minimizing a total loss with respect to target outputs $\target_t$ at
each time,
\begin{equation}
    \mathcal{L}_T = \sum\limits_{t=1}^T \ell_t = \sum\limits_{t=1}^T\ell(s_t, \target_t).
\end{equation}
A typical case is that of a standard recurrent neural network (RNN). In this
case, $s_t = (o_t, h_t)$, where $o_t$ are the activations of the output
layer (encoding the predictions), and $h_t$ are the activations of the hidden recurrent
layer. For this simple RNN, the
dynamical system takes the form
\begin{align}
    h_{t+1} &= \tanh(W_x\,x_{t+1} + W_h\,h_t + b)\\
    o_{t+1} &= W_o h_{t+1}\\
    \ell_{t+1} &= \ell(o_{t+1}, \target_{t+1})
\end{align}
with parameters $\theta = (W_x, W_h, b)$.

Commonly, $\theta$ is optimized via a gradient descent procedure, i.e. iterating 
\begin{equation}
    \theta \leftarrow \theta - \eta \frac{\partial \mathcal{L}_T}{\partial \theta}
\end{equation}
where $\eta$ is the learning rate.
The focus is then to efficiently compute $\partial \mathcal{L}_T/\partial \theta$.

\emph{Backpropagation through time} is a method of choice to perform this computation.
BPTT computes the gradient by unfolding the dynamical system through time and
backpropagating through it, with each timestep corresponding to a layer.
BPTT decomposes the gradient as a sum, over timesteps $t$, of the effect of a change of
parameter at time $t$ on all subsequent losses. Formally,
\begin{equation}
    \frac{\partial \mathcal{L}_T}{\partial \theta} = \sum\limits_{t=1}^T
    \delta \ell_t \, \frac{\partial F}{\partial \theta}(x_t, s_{t-1}, \theta)
    \label{eq:gradient}
\end{equation}
where $\delta \ell_t\deq \frac{\partial \mathcal{L}_T}{\partial s_t}$ is
computed backward iteratively according to the backpropagation equation
\begin{equation}
\begin{dcases}
    \delta \ell_T = \frac{\partial \ell}{\partial s}(s_T, \target_T)\\
    \delta \ell_{t} = \delta \ell_{t+1}\,\frac{\partial F}{\partial s}(x_{t+1}, s_t, \theta) + \frac{\partial \ell}{\partial s}(s_t, \target_t).
    \label{eq:BPTT}
    \end{dcases}
\end{equation}
These backpropagation equations extend the classical ones \cite{jaeger2002tutorial},
which deal with the case of a simple RNN for $F$.
%\NDY{Of all my books, not a single one gives the general BPTT equations...}

Unfortunately, 
BPTT requires processing the full sequence both forward and
backward. This requires maintaining the full unfolded network, or equivalently
storing the full history of inputs and activations (though see
\cite{constant-mem-bptt}). This is impractical when very long
sequences are processed with large networks: processing the whole
sequence at every gradient step slows down learning.

Practically, this is
alleviated
by truncating gradient flows after a fixed number of timesteps, or equivalently, splitting
the input sequence into subsequences of fixed length, and only backpropagating through those
subsequences.
\footnote{Usually the internal state $s_t$ is maintained from
one subsequence to the other, not reset to a default value.}
This algorithm is referred to as \emph{Truncated BPTT}. 
With truncation length $L< T$, the corresponding equations just drop
the recurrent term $\delta {\ell}_{t+1}\,\frac{\partial F}{\partial
        s}(x_{t+1}, s_t, \theta)$ every $L$ time steps, namely,
\begin{align}
    \delta \hat{\ell}_t &\deq  \begin{dcases}
        \frac{\partial \ell}{\partial s}(s_t,
	\target_t) &\text{if $t$ is a multiple of $L$}\\
        \delta \hat{\ell}_{t+1}\,\frac{\partial F}{\partial
	s}(x_{t+1}, s_t, \theta) + \frac{\partial \ell}{\partial s}(s_t,
	\target_t) & \text{otherwise.}
    \end{dcases}
\end{align}
This also allows for online application: for instance, the gradient
estimate from the first subsequence $t=1\ldots,L$ does not depend on
anything at time $t>L$.

However, this gradient estimation scheme is heuristic and provides biased
gradient estimates.
In general
the resulting gradient estimate can be quite far from the true
gradient even with large truncations $L$ (Section~\ref{sec:exp}).
Undesired
behavior, and, sometimes, divergence can follow when performing gradient
descent with truncated BPTT
(Fig.~\ref{fig:bpttf}).

\section{Anticipated Reweighted Backpropagation Through Time:
unbiasedness through reweighted stochastic truncation lengths}
\label{sec:artbp}

% This section introduces ARTBP. ARTBP is designed in the same spirit
% as truncated BPTT,
% but, contrary to the latter, provides provides an unbiased estimate of the gradient.
% 
% \subsection{Unbiasedness through stochastic truncation length}

\begin{figure}
    \centering
    \hspace{-10em}
    \begin{subfigure}{0.40\textwidth}
        \centering
        \includegraphics[width=5cm]{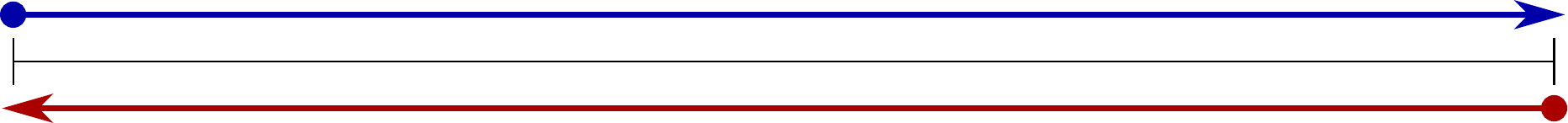}
        \caption{BPTT}
    \end{subfigure}
    \begin{subfigure}{0.40\textwidth}
        \centering
        \includegraphics[width=5cm]{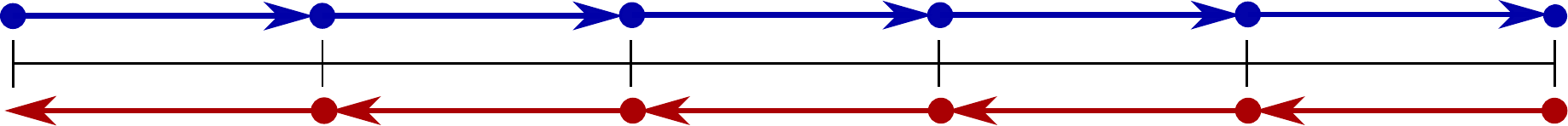}
        \caption{Truncated BPTT}
    \end{subfigure}
    \begin{subfigure}{0.40\textwidth}
        \centering
        \includegraphics[width=5cm]{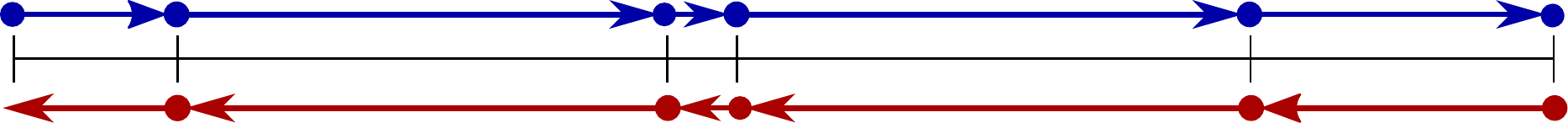}
        \caption{ARTBP}
    \label{fig:artintuition}
    \end{subfigure}
    \hspace{-10em}
    \caption{Graphical representation of BPTT, truncated BPTT and ARTBP. Blue arrows
    represent forward propagations, red arrows backpropagations. Dots represent either
    internal state resetting or gradient resetting.}
    \label{fig:intuition}
\end{figure}

Like 
truncated BPTT, ARTBP splits the initial sequence into subsequences, and
only performs backpropagation through time on subsequences. However, contrary
to the latter, it does not split the sequence evenly. The length of each
subsequence is sampled according to a specific probability distribution. Then the
backpropagation equation is modified by introducing a suitable
reweighting factor at every step to ensure unbiasedness. Figure~\ref{fig:intuition}
demonstrates the difference between BPTT, truncated BPTT and ARTBP.

Simply sampling arbitrarily long truncation lengths does not provide
unbiasedness. Intuitively, it still favors short term gradient terms over
long term ones. When using full BPTT, gradient computations flow back \footnote{
Gradient flows between timesteps $t$ and $t'$ if there are no truncations
occuring between $t$ and $t'$.  } from every timestep $t$ to every 
timestep $t'<t$. In truncated BPTT, gradients do not flow from $t$ to
$t'$ if $t-t'$ exceeds the truncation length. In ARTBP, since random truncations are introduced,
gradient computations flow from $t$ to $t'$ with a certain probability, decreasing
with $t-t'$. To restore balance, ARTBP rescales gradient flows by
their inverse probability.  Informally, if a flow has a probability $p$
to occur, multiplication of the flow by $\frac{1}{p}$ restores balance on
average.

Formally, at each training epoch, ARTBP starts by sampling a random sequence of
truncation points, that is $(X_t)_{1\leq t \leq T} \in \{0, 1\}^T$. A
truncation will occur at all points $t$ such that $X_t=1$.
Here $X_t$ may have a probability law that depends on $X_1, \ldots, X_{t-1}$, and also on
the sequence of states $(s_t)_{1\leq t\leq T}$ of the system.  
%ARTBP
%performs truncated backpropagation on the resulting decomposition of the
%initial sequence into subsequences. 
The reweighting factors that ARTBP introduces in the backpropagation
equation depend on these truncation probabilities.
%ARTBP additionally introduces
%reweighting factors, to obtain unbiasedness. These reweighting factors
%depend on the truncation probabilities.
(Unbiasedness is not obtained just by global importance reweighting between the
various truncated subsequences: indeed, the backpropagation equation
inside each subsequence has to be modified at every time step, see
\eqref{eq:artbp}.)

The question of how to choose good probability distributions for the
truncation points $X_t$ is postponed till Section~\ref{sec:cchoice}.
Actually, 
unbiasedness holds for any choice of truncation probabilities
(Prop~\ref{prop:artbp}), but different choices for $X_t$ lead to
different variances for the resulting gradient estimates.

% Thus, let us introduce random truncations with
% probabilities $c_t$:
% \begin{align}
%     X_t&\deq \begin{cases}
%     1 & \text{if there is a truncation between $t$ and $t+1$}
%     \\ 0 &\text{otherwise}
%     \end{cases}
%   \\
%     c_t &\deq \mathbb{P}(X_t=1\mid X_{t-1}, \ldots, X_1)
% \end{align}
% For now we leave $c_t$ unspecified;
% the choice of $c_t$ is discussed in Section~\ref{sec:cchoice}.
% 
% ARTBP replaces the BPTT equation \eqref{eq:BPTT} by truncating if
% $X_t=1$, and introducing a factor $\frac{1}{1-c_t}$ when no truncation
% occurs:
% \begin{align}
% \label{eq:artbp}
%     \delta \tilde{\ell}_t &\deq \begin{cases}
%         \displaystyle\frac{\partial \ell}{\partial s}(s_t, \target_t) &\text{ if $X_t=1$ or $t=T$}\\
%         \\
%         \displaystyle\frac{1}{1-c_t}\, \delta
% 	\tilde{\ell}_{t+1}\,\frac{\partial F}{\partial s}(x_{t+1}, s_t,
% 	\theta) + \frac{\partial \ell}{\partial s}(s_t, \target_t) &\text{ otherwise}
%     \end{cases}
% \end{align}
% The final estimate $\tilde{g}$ of the total gradient $\frac{\partial
% \mathcal{L}_T}{\partial\theta}$
% is obtained by plugging $\delta \tilde{\ell}_t$ into \eqref{eq:gradient}.

\begin{proposition} 
\label{prop:artbp}
Let $(X_t)_{t=1...T}$ be any sequence of binary random variables, chosen
according to probabilities
\begin{equation}
c_t \deq \mathbb{P}(X_t=1\mid X_{t-1}, \ldots, X_1)
\end{equation}
and assume $c_t\neq 1$ for
all $t$.
%\NDY{Do we also condition wrt $s_{1:T}$? I think it works, but
%might be confusing}

Define ARTBP to be backpropagation through time with a truncation
between $t$ and
$t+1$ iff $X_t=1$, and a compensation factor $\frac{1}{1-c_t}$ when
$X_t=0$, namely:
\begin{align}
\label{eq:artbp}
    \delta \tilde{\ell}_t &\deq \begin{dcases}
        \frac{\partial \ell}{\partial s}(s_t, \target_t) &\text{ if $X_t=1$ or $t=T$}\\
        \frac{1}{1-c_t}\, \delta
	\tilde{\ell}_{t+1}\,\frac{\partial F}{\partial s}(x_{t+1}, s_t,
	\theta) + \frac{\partial \ell}{\partial s}(s_t, \target_t) &\text{ otherwise.}
    \end{dcases}
\end{align}

Let $\tilde{g}$ be the gradient
estimate obtained by using $\delta\tilde{\ell}_t$ instead of
$\delta\ell_t$ in ordinary BPTT \eqref{eq:gradient},
namely
\begin{equation}
\tilde{g}\deq 
     \sum\limits_{t=1}^T \delta \tilde{\ell}_t \, \frac{\partial
     F}{\partial \theta}(x_t, s_{t-1}, \theta)
\end{equation}
Then, on average over the ARTBP truncations, this is an unbiased gradient
estimate of the total loss:
    \begin{equation}
    \mathbb{E}_{X_1, \ldots, X_T}\left[\,\tilde{g}\,\right] =
    \frac{\partial \mathcal{L}_T}{\partial \theta}.
    \end{equation}
\end{proposition}

The core of the proof is as follows: With probability $c_t$ (truncation),
$\delta\tilde\ell_{t+1}$ does not contribute to $\delta\tilde\ell_t$.
With probability $1-c_t$ (no truncation), it contributes with a factor
$\frac{1}{1-c_t}$. So on average, $\delta\tilde\ell_{t+1}$ contributes to
$\delta\tilde\ell_t$ with a factor $1$, and ARTBP \eqref{eq:artbp}
reduces to standard, non-truncated BPTT \eqref{eq:BPTT} on average. The
detailed proof is given in Section~\ref{sec:proof}.

While the ARTBP gradient estimate above is unbiased, some noise is
introduced due to stochasticity of the truncation points. It turns out
that ARTBP trades off memory consumption (larger truncation lengths) for
variance, as we now discuss.

\section{Choice of $c_t$ and memory/variance tradeoff}
\label{sec:cchoice}

ARTBP requires specifying the probability $c_t$ of truncating at time
$t$ given previous truncations.
Intuitively the $c$'s regulate the
average truncation lengths.  For instance, with a constant $c_t\equiv c$, the
lengths of the subsequences between two truncations follow a geometric
distribution, with average truncation length
$\frac{1}{c}$.  Truncated BPTT with fixed truncation length $L$ and ARTBP with fixed
$c=\frac{1}{L}$ are thus comparable memorywise.

Small values of $c_t$ will lead to long subsequences and gradients closer
to the exact value, while large values
will lead to shorter subsequences but larger compensation factors
$\frac{1}{1-c_t}$ and noisier estimates. In particular, the product of
the $\frac{1}{1-c_t}$ factors inside a subsequence can grow quickly. For
instance, a constant $c_t$ leads to exponential growth of the
cumulated $\frac{1}{1-c_t}$ factors when iterating \eqref{eq:artbp}.
% In parallel, $c$'s also regulate the variance of the gradient approximation.
% Considering extreme cases, when the sequence of $c$'s is constant, equal to $1$, no truncation
% happens, and the estimate provided by ARTBP is the exact gradient.
% Generally, when the sequence of $c$'s is a non unitary constant, the probability of
% a truncation of length $T$ decreases exponentially with $T$. The flow reweighting factor
% thus increases exponentially with the length of the truncation, making the approximation
% extremely variant.

To mitigate this effect, we suggest to set $c_t$ to values such that the
probability to have a subsequence of length $L$ decreases like
$L^{-\alpha}$. The variance of the lengths of the subsequences will be
finite if $\alpha>3$. Moreover we might want to control the average truncation length $L_0$. This is achieved via
\begin{equation}
    c_t=\mathbb{P}(X_t=1\mid X_{t-1}, \ldots, X_1) =
    \frac{\alpha-1}{(\alpha-2) L_0 + \delta t}
    \label{eq:ct}
\end{equation}
where $\delta t$ is the time elapsed since the last truncation,
$\delta t=t-\sup \{s\mid s<t, X_s=1\}$. % (which only depends on $X_{t-1}, \ldots, X_1$).
%, and $L_0$ and $\alpha$ are parameters to be set.
Intuitively, the more time spent without truncating, the lower the
probability to truncate. This formula is chosen such that the average
truncation length is approximately $L_0$, and the standard deviation
from this average length is finite. The parameter $\alpha$ controls the
regularity of the distribution of truncation lengths: all moments lower
than
$\alpha-1$ are finite, the others are infinite. With larger $\alpha$,
large lengths will be less frequent, but the compensating factors
$\frac{1}{1-c_t}$ will be larger.

With this choice of $c_t$, the product of the $\frac1{1-c_t}$ factors
incurred by backpropagation inside each subsequence grows polynomially
like $L^{\alpha-1}$ in a subsequence of length $L$. If the dynamical system
has geometrically decaying memory, i.e., if the operator norm of the transition
operator $\frac{\partial F}{\partial s}$ is less than $1-\epsilon$ most of
the time, then the
value of $\delta\tilde\ell_t$ will stay controlled, since
$(1-\epsilon)^L\cdot L^\alpha$ stays bounded. On the other hand, using a
constant $c_t\equiv c$ provides bounded $\delta\tilde\ell_t$ only for small
values $c<\epsilon$.

In the experiments below, we use the $c_t$ from \eqref{eq:ct} with
$\alpha=4$ or $\alpha=6$.

% Fig.~\ref{fig:hist} displays the distribution of truncation lengths for different value of
% $\alpha$, with the same $t_0$.
% \begin{figure}
%     \centering
%     \begin{subfigure}{0.32\textwidth}
%         \centering
%         \scalebox{0.38}{\input{plots/hist/hist-3.tex}}
%     \end{subfigure}
%     \begin{subfigure}{0.32\textwidth}
%         \centering
%         \scalebox{0.38}{\input{plots/hist/hist-5.tex}}
%     \end{subfigure}
%     \begin{subfigure}{0.32\textwidth}
%         \centering
%         \scalebox{0.38}{\input{plots/hist/hist-10.tex}}
%     \end{subfigure}
%     \caption{Histograms of truncation lengths. From left to right, $\alpha=4,
%     6, 11$\NDY{I shifted $\alpha$ by $1$...}. Variance is reduced as $\alpha$ is increased.}
%     \label{fig:hist}
% \end{figure}

\section{Online implementation}

Importantly, ARTBP can be directly applied online, thus providing unbiased gradient
estimates for recurrent networks.

Indeed, not all truncation
points have to be drawn in advance: ARTBP can be applied  by
sampling the first truncation point, performing both forward and
backward passes of BPTT up until this point, and applying a partial
gradient descent update based on the resulting gradient on this
subsequence. Then one moves to the next subsequence and the next
truncation point, etc.\ (Fig.~\ref{fig:artintuition}).

%Pseudo-code for online application of ARTBP is provided as supplementary material.

\section{Experimental validation}
\label{sec:exp}

The experimental setup below aims both at illustrating
the theoretical properties of ARTBP compared to truncated BPTT,
and at testing the soundness of ARTBP on real world data.

\subsection{Influence balancing}

The influence balancing experiment is a synthetic example 
demonstrating, in a very simple model, the importance of being unbiased. 
Intuitively, a parameter has a positive short term influence, but a
negative long term one that surpasses the short term effect. Practically,
we consider a row of agents, numbered from left to
right from $1$ to $p+n$ who, at each time step, are provided with a
signal depending on the parameter, and diffuse part of their current
state to the agent directly to their left.  The $p$ leftmost agents
receive a positive signal at each time step, and the $n$ rightmost agents
a negative signal. The training goal is to control the state of the
leftmost agent.  The first $p$ agents contribute positively to the first
agent state, while the next $n$ contribute negatively.  However, agent
$1$ only feels the contribution from agent $k$ after $k$ timesteps. If
optimization is blind to dependencies above $k$, the effect of $k$ is never
felt. A typical instantiation of such a problem would be that of a drug
whose effect varies after various delays; the parameter to be optimized is
the quantity of drug to be used daily.

\begin{figure}
    \centering
    \includegraphics[width=7cm]{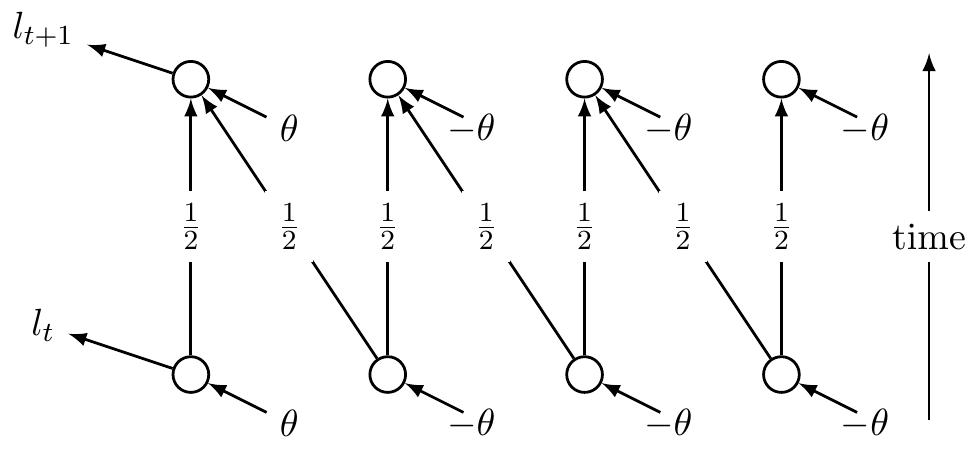}
    \caption{Influence balancing dynamics, $1$ positive influence, $3$ negative influences.}
    \label{fig:infl-bal}
\end{figure}

Such a model can be formalized as \cite{uoro}
\begin{equation}
    s_{t+1} = A \, s_t + (\theta, \ldots, \theta, -\theta, \ldots, -\theta)\trans
\end{equation}
with $A$ a square matrix of size $p+n$ with $A_{k,k} = 1/2$, $A_{k,k+1}=1/2$, and
$0$ elsewhere; $s_t^k$ corresponds to the state of the $k$-th agent. $\theta
\in \real$ is a scalar parameter corresponding to the intensity of the signal
observed at each time step. 
The right-hand-side has $p$ positive-$\theta$ entries and $n$ negative-$\theta$ entries.
The loss considered is an arbitrary target on the leftmost agent $s^1$,
\begin{equation}
    \ell_t = {\textstyle \frac12} (s^1_t - 1)^2.
\end{equation}
The dynamics is illustrated schematically in Figure~\ref{fig:infl-bal}.

Fixed-truncation BPTT is experimentally compared with
ARTBP for this problem. The setting is online: starting at $t=1$, a first truncation length $L$ is selected (fixed for BPTT, variable
for ARTBP), forward and backward passes are performed on the subsequence
$t=1,\ldots,L$, a vanilla gradient step is performed with the resulting
gradient estimate, then the procedure is repeated with the next
subsequence starting at $t=L+1$, etc..

Our experiment uses $p=10$ and $n=13$, so that after $23$ steps the signal
should have had time to travel through the network.
Truncated BPTT is tested with various truncations $L=10,100,200$. 
(As the initial $\theta$ is fixed, truncated BPTT is deterministic in
this experiment, thus we only provide a single run for each $L$.) ARTBP is tested
with the probabilities \eqref{eq:ct} using $L_0=16$ (average truncation
length) and $\alpha=6$. ARTBP is stochastic: five random runs are
provided to test reliability of convergence. 

\begin{figure}[h!]
    \centering
    \scalebox{0.5}{\input{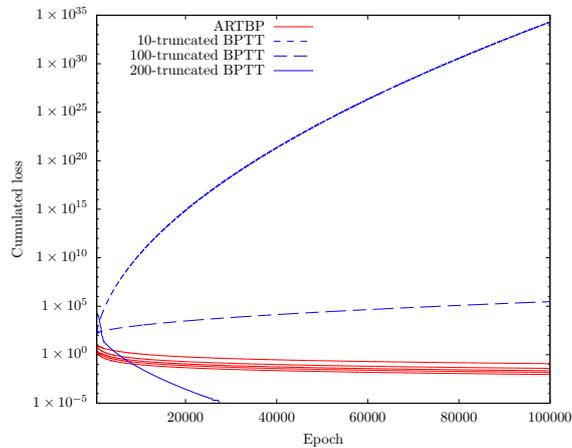}}
    \caption{ARTBP and truncated BPTT on influence balancing, $n=13$, $p=10$. Note the log scale
    on the $y$-axis.}
    \label{fig:bpttf}
\end{figure}

The results are displayed in Fig.~\ref{fig:bpttf}.
We used decreasing learning rates $\eta_t = \frac{\eta_0}{\sqrt{1 + t}}$
where $\eta_0=3\times 10^{-4}$ is the initial learning rate and $t$ is
the timestep.
We plot the average loss over timesteps $1$ to $t$, as a function of $t$.

Truncated BPTT diverges even for truncation ranges largely above the
intrinsic temporal scale of the system. This is an
expected result: due to bias, truncated BPTT ill-balances
temporal dependencies and estimates the overall gradient with a wrong sign. In
particular, reducing the learning rate will \emph{not} prevent divergence. On the
other hand, ARTBP reliably converges on every run.

Note that for the largest truncation $L=200$, truncated BPTT finally
converges, and does so at a faster rate than ARTBP. This is because this
particular problem is deterministic, so that a deterministic gradient scheme will
converge (if it does converge) geometrically like $O(e^{-\lambda t})$, whereas ARTBP is stochastic
due to randomization of truncations, and so will not converge faster than
$O(t^{-1/2})$. This difference would disappear, for instance, with noisy
targets or a noisy system.%\NDY{add N(0,1) to the model?}

\paragraph{Character-level Penn Treebank language model.}
We compare ARTBP to truncated BPTT on the character-level version of
the Penn Treebank dataset, a standard set of case-insensitive,
punctuation-free English text \cite{ptb}. Character-level language modelling
is a common benchmark for recurrent models.

\begin{figure}
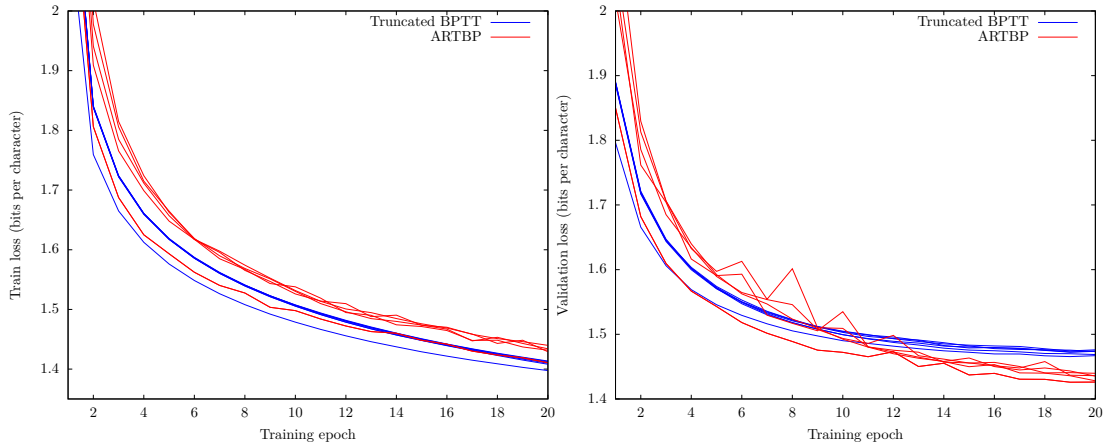

    \hspace{-1cm}
    \begin{subfigure}{0.47\textwidth}
        \centering
        \scalebox{0.5}{\input{ptb_train-arxiv.tex}}
        \caption{Learning curves on Penn Treebank train set.}
    \end{subfigure}
    \hspace{1cm}
    \begin{subfigure}{0.47\textwidth}
        \centering
        \scalebox{0.5}{\input{ptb-arxiv.tex}}
        \caption{Learning curves on Penn Treebank validation set.}
    \end{subfigure}
    \hspace{-1cm}
%    \hspace{1.5cm}
%    \begin{subtable}{0.54\textwidth}
%        \centering
%        \begin{tabular}{l c}
%            \toprule
%            \bf{Model} & \bf{Test loss (BPC)} \\
%            \midrule
%            LSTM Truncated BPTT (ours) & $1.43$ \\
%            LSTM ARTBP (ours) & $1.40$ \\
%            \midrule
%            LSTM \cite{bn_bench} & $1.38$ \\
%            BN-LSTM \cite{bn_bench} & $1.32$ \\
%            Zoneout \cite{zoneout_bench} & $1.27$ \\
%            LSTM \cite{graves_bench} & $1.26$ \\
%            HM-LSTM \cite{hm_bench} & $1.24$ \\
%            HyperNetworks \cite{hn_bench} & $\pmb{1.22}$\\
%            \bottomrule
%        \end{tabular}
%	\caption{Compression rate on Penn Treebank test set.}
%	\label{fig:pennsota}
%    \end{subtable}
    \caption{Results on Penn Treebank character-level language modelling.}
    \label{fig:ptb}
\end{figure}

The dataset is split into training, validation and test sets
following
\cite{ptb_proc}. Both ARTBP and truncated BPTT are used to train an LSTM model \cite{lstm}
with a softmax classifier on its hidden state, on the character prediction task. 
The training set is batched into $64$ subsets processed in parallel to
increase computing speed.
Before each full pass on the training set, the batched training sequences are split into
subsequences:
\begin{itemize}[noitemsep]
    \item for truncated BPTT, of fixed size $50$;
    \item for ARTBP, at random following the scheme \eqref{eq:ct}
	with $\alpha=4$ and $L_0=50$.
\end{itemize} 
Truncated BPTT and ARTBP process these subsequences sequentially,
\footnote{
    Subsequences are not shuffled, as we do not reset the internal state of the network
    between subsequences.
} as in
Fig.~\ref{fig:intuition}. The parameter is updated
after each subsequence, using the Adam \cite{adam} stochastic gradient
scheme, 
with learning rate $10^{-4}$. The biases of the LSTM unit forget
gates are set to $2$, to prevent early vanishing gradients
\cite{forget_init}. Results (in bits per character, bpc) are displayed
in Fig.~\ref{fig:ptb}. Six randomly sampled runs are plotted, to test
reliability.

In this test, ARTBP slightly outperforms truncated BPTT in terms of
validation and test error, while the reverse is true for the training error
(Fig.~\ref{fig:ptb}).

Even with ordinary truncated BPTT, we could not reproduce reported state of the art
results, and do somewhat worse. We reach a test error of
$1.43$ bpc with standard truncated BPTT and $1.40$ bpc with ARTBP, while
reported values with similar LSTM models range from $1.38$ bpc
\cite{bn_bench}
to $1.26$ bpc \cite{graves_bench} (the latter with a different test/train
split).
This may be due to differences in the experimental setup: we have applied
truncated BPTT without subsequence shuffling or gradient
clipping \cite{graves_bench} (incidentally, both would break unbiasedness).
%Shuffling of the learning sequences is not
%applicable in our case, as we need to maintain the internal state of the
%network between sequences\NDY{unclear. If they reset the internal state
%we could do it too...}. Gradient
%clipping \cite{graves_bench} would break unbiasedness. 
Arguably,
the numerical issues solved by gradient clipping are model specific, not
algorithm specific, while the point here was to compare ARTBP to
truncated BPTT for a given model.

\section{Conclusion}

We have shown that the bias introduced by truncation in the
backpropagation through time algorithm can be compensated by the simple
mathematical trick of randomizing the truncation points and introducing
compensation factors in the backpropagation equation. The algorithm is
experimentally viable, and provides proper balancing of the effects of
different time scales when training recurrent models.

\section{Proof of Proposition~\ref{prop:artbp}}
\label{sec:proof}

First, by backward induction, we show that for all $t\leq T$, for all
$x_1,\ldots,x_{t-1}\in\{0,1\}$,
\begin{equation}
\label{eq:induction}
\E \left[ \delta\tilde\ell_t \mid X_{1:t-1}=x_{1:t-1} \right] =
\delta\ell_t
\end{equation}
where $\delta\ell_t$ is the value obtained by ordinary BPTT
\eqref{eq:BPTT}. Here $x_{1:k}$ is short for $(x_1,\ldots,x_k)$.

For $t=T$, this holds by definition: $\delta \tilde{\ell}_T =
\frac{\partial \ell}{\partial s}(s_T, \target_T) = \delta \ell_T$.

Assume that the induction hypothesis \eqref{eq:induction} holds at time $t+1$.
Note that the values
$s_t$ do not depend on the random variables $X_t$, as they are computed
during the forward pass of the algorithm. In particular, the various
derivatives of $F$ and $\ell$ in \eqref{eq:artbp} do not depend on
$X_{1:T}$.
% Moreover, by \eqref{eq:artbp}, $\delta\tilde\ell_t$ is computed using
% $X_t,\ldots,X_T$ but not $X_{t'}$ for $t'<t$. Therefore, if the random
% variables $X_t$ are all independent, then $\delta\tilde\ell_{t+1}$ is
% independent from $X_t$.

Thus
\begin{align}
\E & \left[ \delta\tilde\ell_t \mid X_{1:t-1}=x_{1:t-1}
\right]=\nonumber\\
&
%\E \left[ 
\Pr(X_t=1\mid X_{1:t-1}=x_{1:t-1}) \,\E \left[
\delta\tilde\ell_t \mid X_{1:t-1}=x_{1:t-1}, X_t=1 \right]
+%\right.\nonumber
\\&%\left.
\quad
\Pr(X_t=0\mid X_{1:t-1}=x_{1:t-1}) \,\E \left[
\delta\tilde\ell_t \mid X_{1:t-1}=x_{1:t-1},X_t=0 \right]
%\right]
\\
=&\;%\E \left[ 
c_t \,\E \left[
\delta\tilde\ell_t \mid X_{1:t-1}=x_{1:t-1},X_t=1 \right]
+(1-c_t)\,
\E \left[
\delta\tilde\ell_t \mid X_{1:t-1}=x_{1:t-1},X_t=0
\right]%\right]
\label{eq:conditioned}
\end{align}

If $X_t=1$ then $\delta\tilde\ell_t=\frac{\partial \ell}{\partial s}
(s_t, \target_t)$. If $X_t=0$, then $\delta\tilde\ell_t=
                \frac{\partial \ell}{\partial s} (s_t, \target_t)
                + \frac{1}{1-c_t}\,\delta
		\tilde{\ell}_{t+1}\,\frac{\partial F}{\partial
		s}(x_{t+1}, s_t, \theta)$.
Therefore, substituting into \eqref{eq:conditioned},
\begin{align}
\E & \left[ \delta\tilde\ell_t \mid X_{1:t-1}=x_{1:t-1}\right]
=%\E \left[ 
\frac{\partial \ell}{\partial s} (s_t, \target_t)
                +
		\E \left[ \delta\tilde{\ell}_{t+1} \mid
		X_{1:t-1}=x_{1:t-1},X_t=0  \right]
		\frac{\partial F}{\partial
		                s}(x_{t+1}, s_t, \theta)
%\right]
\end{align}
but by the induction hypothesis at time $t+1$, this is exactly
$\frac{\partial \ell}{\partial s} (s_t,
\target_t)+\delta\ell_{t+1}\frac{\partial F}{\partial s}(x_{t+1}, s_t,
\theta)$, which is $\delta\ell_t$.

Therefore, $\E\left[\delta\tilde\ell_t\right]=\delta\ell_t$
unconditionally.
Plugging the $\delta \tilde{\ell}$'s into \eqref{eq:gradient}, and averaging
    \begin{align}
        \mathbb{E}_{X_1, \ldots, X_T} \left[\,\tilde{g}\,\right]
        &=
        \sum\limits_{t=1}^T \mathbb{E}_{X_t, \ldots, X_T} \left[\delta
	\tilde{\ell}_t\right]\frac{\partial F}{\partial \theta}(x_{t},
	s_{t-1}, \theta)\\
        &=
        \sum\limits_{t=1}^T  \delta \ell_t\,\frac{\partial F}{\partial
	\theta}(x_{t}, s_{t-1}, \theta)\\
        &= \frac{\partial \mathcal{L}_T}{\partial \theta}
    \end{align}
    which ends the proof.

\bibliographystyle{alpha}
\bibliography{artbp}
\end{document}